\documentclass[letterpaper, 10 pt, conference]{ieeeconf}  
\usepackage[utf8]{inputenc}
\usepackage{graphicx}
\usepackage{multirow}
\usepackage{float}

\IEEEoverridecommandlockouts                              

\usepackage{amsmath} 
\usepackage{comment}
\usepackage{algorithm}
\usepackage{algpseudocode}

\title{\LARGE \bf
Zero-Shot Parameter Learning of Robot Dynamics Using\\ Bayesian Statistics and Prior Knowledge
}

\author{Carsten Reiners$^{1,*}$, Minh Trinh$^{1,*,**}$, Lukas Gründel$^{1}$, Sven Tauchmann$^{2}$,  \\David Bitterolf$^{3}$, Oliver Petrovic$^{1}$ and Christian Brecher$^{1}$
\thanks{*Equal contribution}
\thanks{$^{1}$Laboratory of Machine Tools and Production Engineering of RWTH Aachen University, Aachen, Germany}%
\thanks{$^{2}$Business Unit Motion Control of Siemens AG, Chemnitz, Germany}%
\thanks{$^{3}$Business Unit Motion Control of Siemens AG, Erlangen, Germany}%
\thanks{$^{**}$Corresponding author
        {\tt\small m.trinh@wzl.rwth-aachen.de}}%
}

\begin{document}

\maketitle
\thispagestyle{empty}
\pagestyle{empty}

\begin{abstract}

Inertial parameter identification of industrial robots is an established process, but standard methods using Least Squares or Machine Learning do not consider prior information about the robot and require extensive measurements.
Inspired by Bayesian statistics, this paper presents an identification method with improved generalization that incorporates prior knowledge and is able to learn with only a few or without additional measurements (Zero-Shot Learning). 
Furthermore, our method is able to correctly learn not only the inertial but also the mechanical and base parameters of the MABI Max 100 robot while ensuring physical feasibility and specifying the confidence intervals of the results. 
We also provide different types of priors for serial robots with 6 degrees of freedom, where datasheets or CAD models are not available.

\end{abstract}

\section{INTRODUCTION AND BACKGROUND}

In recent years, industrial robots have been used in manufacturing applications that require higher accuracies of position and path than conventional handling or welding tasks \cite{VERL2019799}. These applications include machining, laser cutting or additive manufacturing. In order to ensure the precision of the robotic system, model-based control algorithms can be used to predict and compensate deviations of the tool center point (TCP) \cite{grundel2024modellierungsmethoden}. The accuracy of the used models highly depends on the inertial parameters such as masses or the inertia tensors of each robot link. In many cases, CAD models lack availability or accuracy for real robot systems, which also include unmodeled attachments, such as customized end effectors \cite{swevers2007dynamic}. Therefore, many methods for inertial parameter identification have been developed, which are mainly based on Least Squares (LS), or more recently, Machine Learning (ML) methods \cite{gautier2013identification, app11094303}.
LS approaches use a simple and efficient solution to model inertial dynamics. After formulating the inverse dynamics as:
\begin{equation}
  \boldsymbol{\tau}(\boldsymbol{q}, \dot{\boldsymbol{q}}, \ddot{\boldsymbol{q}})
  \;=\;
  \mathbf{M}(\boldsymbol{q}) \,\ddot{\boldsymbol{q}}
  \;+\;
  \mathbf{C}(\boldsymbol{q}, \dot{\boldsymbol{q}})\,\dot{\boldsymbol{q}}
  \;+\;
  \mathbf{g}(\boldsymbol{q}),
  \label{eq:invdyn}
\end{equation}
with the joint positions, velocities and accelerations $\boldsymbol{q}, \dot{\boldsymbol{q}}$ and $\ddot{\boldsymbol{q}}$. $\boldsymbol{\tau}(\boldsymbol{q}, \dot{\boldsymbol{q}}, \ddot{\boldsymbol{q}})$ is the torque, $\mathbf{M}(\boldsymbol{q})$ the mass matrix, $\mathbf{C}(\boldsymbol{q}, \dot{\boldsymbol{q}})$ the Coriolis and centripetal forces, and $\mathbf{g}(\boldsymbol{q})$ the vector of gravitational torques.
The linear dependencies are realized and the equation is defined using a regressor $\mathbf{W}(\boldsymbol{q},\dot{\boldsymbol{q}},\ddot{\boldsymbol{q}})$ and a set of independent base parameters (BP) $\boldsymbol{\Phi}$ \cite{619069}:
\begin{equation}
  \boldsymbol{\tau} = \mathbf{W}(\boldsymbol{q},\dot{\boldsymbol{q}},\ddot{\boldsymbol{q}})\,\boldsymbol{\Phi},
  \label{eq:regressor}
\end{equation}
Since LS approaches do not guarantee identifiability and physical consistency of the BP, Gautier eliminated parameters of high variance and modified a classical weighted least squares approach to select a BP solution close to a given CAD model \cite{gautier2013identification}.
LS and ML approaches provide good models for the measured output, but cannot provide the interpretable mechanical parameters (MP) such as  individual masses necessary for classical mechanical calculations like the recursive Newton Euler algorithm (RNEA) or multibody dynamics simulation. 
As a conclusion, these approaches are only unidirectional, neglect any type of prior knowledge and require extensive measurements.

\cite{gaz2016extracting} suggested optimization-based solutions in the MP space to learn individual CAD-like parameters.
Through lower and upper limits on individual parameters, physical feasibility is ensured. With the problem being underdetermined, the definition of the optimization space and the bounds could lead to marginally feasible solutions \cite{lee2019geometric}. Despite using CAD models to inform constraints, approaches based on linear matrix inequalities (LMI) still result in divergent solutions, even for the same robot type \cite{sturz2017parameter, xu2020dynamic}.
Furthermore, these models cannot quantify the uncertainties for the resulting parameters.
Here, statistical approaches can be effective. Starting from linear dynamic system identification \cite{blanco2015multibody} or state estimation \cite{pillonetto2023full}, especially strategies such as Gaussian process regression (GPR) have been successfully used to combine a kernelized rigid body dynamics (RBD) model with a Gaussian uncertainty model \cite{nguyen2010using}.

To establish a robust robot model, we focus on the RBD and present a new method based on Bayesian statistics for learning base, mechanical, and inertial parameters of industrial robots. Our method uses prior information, ensures physically plausible results, and specifies uncertainties as confidence intervals. 
Here, prior knowledge can vary from diffuse to informative depending on the information available to the user, such as datasheets or CAD models.
Furthermore, not only hard, but also soft boundary constraints on parameters can be introduced to regularize the optimization space. 
We also interpret the prior as pretraining, allowing model outputs without the need for experimental measurements, and thus consider it a Zero-Shot-Learning (ZSL) approach \cite{NIPS2013_2d6cc4b2}. 

\section{PROPOSED METHOD AND EXPERIMENTS}

\begin{figure}[ht]
    \centering
    \includegraphics[scale=0.75]{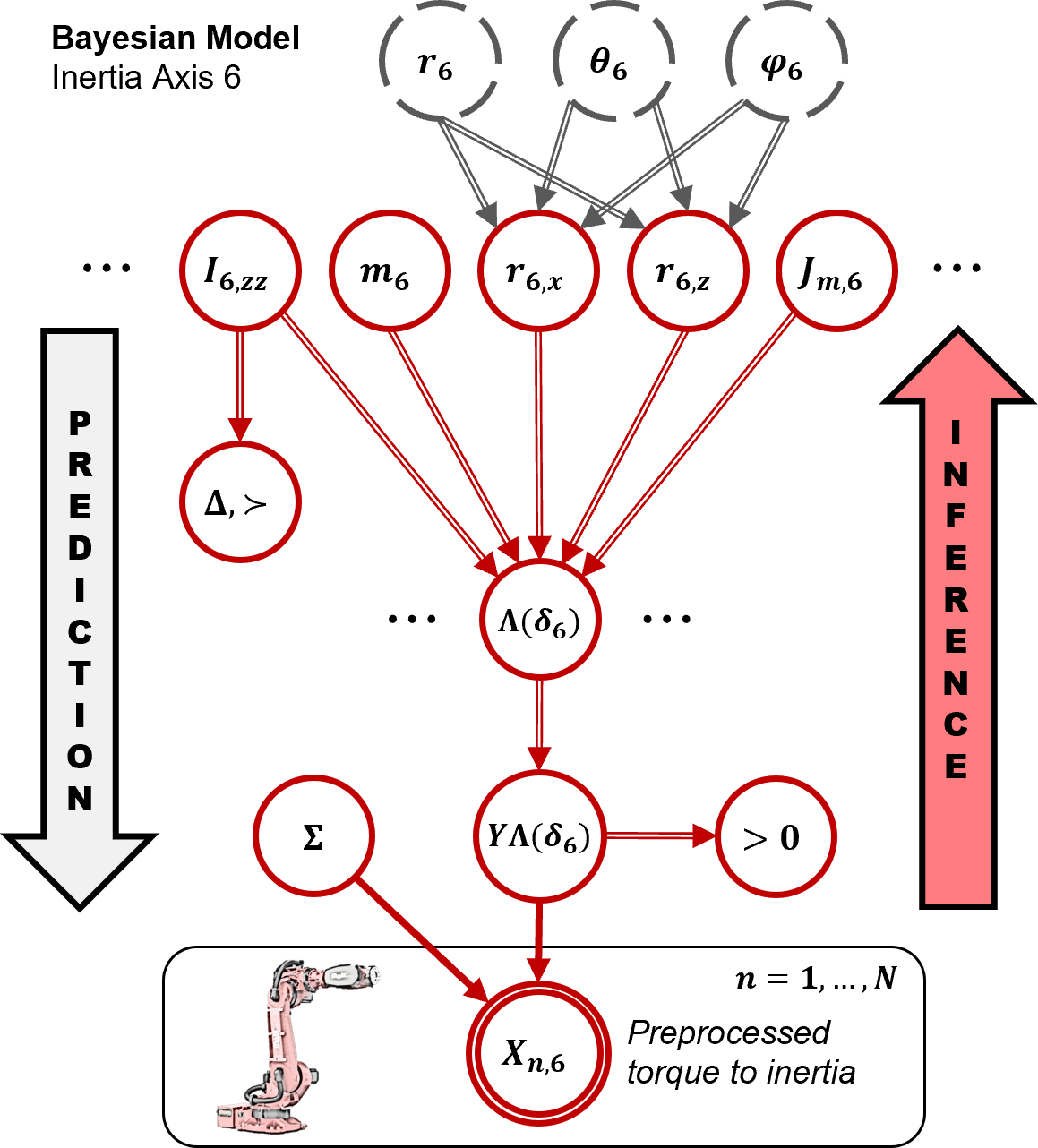}
    \caption{'Bayesian network' exemplified by the inertial model of robot axis 6 with essential nodes colored in red and optional ones in dashed gray. Nodes resemble random variables and double encircled ones represent observed data. Single-lined edges depict statistical and double-lined edges algebraic dependencies. The network has a prediction (Zero-shot) and inference mode.}
    \label{fig:bayes_network}
\end{figure}

Fig. \ref{fig:bayes_network} shows the proposed method via the example of the last axis of the MABI Max 100, an industrial serial robot with six degrees of freedom (DOF). 
The method focuses on the robot's inertial model (Section II.A.), which is represented as a Bayesian network \cite{bishop2006pattern}. 
All mechanical parameters are modeled as random variables to include prior knowledge. Since most relations such as \eqref{eq:regressor} are analytically tractable, they define explicit random variable transformations. Hence, foundations of Bayesian statistics are discussed in (Section II.B.). Then options for including prior knowledge and constraints, which affect multiple parameters, are presented (Section II.C.). 
The Bayesian framework enables multiple operational modes (Section II.D.): prior-based predictions of dynamics may be performed Zero-Shot, possibly augmented by information inferred from constraints, without additional measurements of the robot. In the posterior inference mode, only a small number of measurements are used for learning. Thus, the method is able to learn the mechanical, base and inertial parameters of the dynamic system along with the respective uncertainties.


\subsection{Robot Dynamics and Inertial Parameters}
Our Bayesian method is applied to the identification of an axis-wise inertia model. Given a kinematic description, we seek to identify a set of 11 unknown inertial MP $\boldsymbol{\delta}$ per body $k$. These consist of a mass $m$, the center of mass (CoM) $r$ coordinates, the six independent inertia tensor components of $I$ and the motor inertia $J$. 
\begin{equation}
\resizebox{0.9\hsize}{!}{%
    $\displaystyle
    \boldsymbol{\delta_{k}}=
    \begin{bmatrix}
    m_k\;r_{kx}\;r_{ky}\;r_{kz}\;I_{kxx}\;I_{kxy}\;I_{kxz}\;I_{kyy}\;I_{kyz}\;I_{kzz}\;J_{k}
    \end{bmatrix}
    $%
}
\end{equation}

Of these MP, only those contributing to the inertia-induced joint torque are considered, leading to a new regressor $\mathbf{Y}(\boldsymbol{q})$ with a base parameter vector $\boldsymbol{\Lambda}(\boldsymbol{\delta})$  which models inertia. To investigate the nonlinear dependency of BP on the MP, we will consider an example for axis six:
\begin{equation}
    \Lambda(\boldsymbol{\delta}_6) = J_{6}+m_6r_{6x}^2+m_6r_{6z}^2+I_{6zz}
    \label{eq:BP_AX6}
\end{equation}
With slight abuse of notation, $\Lambda(\boldsymbol{\delta}_6)$ denotes the single base parameter of axis 6 instead of the BP vector. For the complete set of BP for the MABI Max 100 we refer to \cite{grundel2024modellierungsmethoden}.
$\mathbf{X}$ is a set of pose dependent inertia with measurement noise $\boldsymbol{\epsilon}$.
\begin{equation}
    \mathbf{X}=\mathbf{Y}(\boldsymbol{q})\boldsymbol{\Lambda}(\boldsymbol{\delta})+\boldsymbol{\epsilon}
    \label{DefinitionX}
\end{equation}


\subsection{Bayesian Statistics for Learning of Robot Dynamics}

In typical parameter estimation problems, a point estimate for the true value is identified. By changing to a probabilistic framework, we can associate each MP $\delta_i$, BP $\Lambda_m$ or pose-dependent inertia $X_n$ with a random variable and an associated probability distribution. We assume that all MP $\delta_i$ are statistically independent random variables described by a probability density function (PDF) $p_i(\delta_i)$, which represents a distribution $P_i(\pi)$ with parameters $\pi$. 
\begin{equation}
\delta_i \sim P_i(\pi), \quad \delta_i\perp\delta_j,\quad\forall i\neq j
\end{equation}
The deterministic functions $\boldsymbol{\Lambda}(\boldsymbol{q})$ and $\mathbf{Y}(\boldsymbol{q})\boldsymbol{\Lambda}(\boldsymbol{\delta})$ can also be embedded in this framework to numerically derive a direct mapping of the probability distributions for $\boldsymbol{\delta}$ to the distributions of $\Lambda_m$ and $X_n$. As can be seen from the BP example \eqref{eq:BP_AX6} and \eqref{DefinitionX}, only three mathematical operations are required to describe the mappings: sum, product, and scalar multiplication. While a sum of independent random variables $\delta_i$ does not equate to a sum of the PDF, it can be calculated as a convolution integral. Similar formulas for product and scalar multiplication are available \cite{feller1991introduction}.
In the prediction step, our method performs the random variable equivalent operations on the MP PDF to calculate the BP PDF and the pose-dependent inertia without measurement data, using only prior knowledge of the MP as described in the next chapter (Section II.C.), the symbolic BP equations of the robot and the linear relationship between BP and pose-dependent inertia.
In the inference step, the posterior distributions of the MP, BP and pose-dependent inertia are identified, i.e. the probability of parameters given an evidence, in our case the measurement data.
The posterior can be formulated using Bayes' theorem \cite{gelman1995bayesian} via prior $p(\boldsymbol{\delta})$ and likelihood function $p(\mathbf{X} \mid \boldsymbol{\delta})$ given the evidence $p(\mathbf{X})$:
\begin{equation}
    p(\boldsymbol{\delta} \mid \mathbf{X})
\;=\;
\frac{p(\mathbf{X} \mid \boldsymbol{\delta})\; p(\boldsymbol{\delta})}{p(\mathbf{X})}
\label{eq:posterior}
\end{equation}
For \eqref{eq:posterior}, a joint prior probability is necessary:
\begin{equation}
    p(\boldsymbol{\delta})=\prod_ip_i(\delta_i)
\end{equation}
Since the true likelihood function for measurements given the MP is unknown, we assume a Gaussian likelihood:
\begin{equation}
    p(\mathbf{X} \mid \boldsymbol{\delta}) \sim N(\boldsymbol{\mu}, \Sigma)
\end{equation} 
In our case, the mean values $\boldsymbol{\mu}$ are the measured inertias and a diagonal covariance $\boldsymbol{\Sigma}$ is chosen based on the measured value to adapt to different axes and positions. We define the covariance matrix $\Sigma_{ii}=(c\mu_i)^2$ and $c$ as one optimization parameter for all axes during inference. This is because a major source of uncertainty lies in the inaccurate, local identification of the joint stiffness, which we assume to exert the same proportional effect on the inertia measurement of all axes. 
Since the posterior is not directly accessible, inference (sampling from the posterior) is necessary. To do so, the Metropolis algorithm outlined in Algorithm \ref{alg:MH} can be used which also enables us to neglect the factor $p(\mathbf{X})$ in \eqref{eq:posterior} \cite{gelman1995bayesian}.  
\begin{equation}
  p(\boldsymbol{\delta} \mid \mathbf{X})
  \;\propto\;
  p(\boldsymbol{\delta})\;
  p(\mathbf{X}\mid \boldsymbol{\delta}),
  \label{eq:bayes}
\end{equation}
We implement an adaptive Metropolis algorithm in Matlab, where the proposal distribution covariance matrix is continuously updated based on the sample history \cite{haario2001adaptive}. A decrease in sample autocovariance was achieved in our experiments, which is a common quality indicator \cite{geyer1992practical}. Dependent samples reduce the effective sample size, so we ensure that at least 100 effective samples (on average $> 1000$) per parameter are available \cite{Geyer2011-yi}. In addition, \cite{betancourt2017conceptual} motivated a regular optimization space through rescaling or coordinate changes. The coordinate change and the additional parameter $c$ for the covariance model lead to a new sampling space $\boldsymbol{\delta}\mapsto\boldsymbol{\hat{\delta}}$.
Using the samples from the Markov chain, the posterior PDF of all parameters can be calculated.

\begin{algorithm}
\caption{Metropolis Algorithm}
\label{alg:MH}
\begin{algorithmic}[1]
\State \textbf{Input:} Prior density $p(\boldsymbol{\delta})$, Likelihood function $p(\mathbf{X}\mid\boldsymbol{\delta})$, proposal distribution $q(\boldsymbol{\delta}' \mid \boldsymbol{\delta})$, initial state $\boldsymbol{\delta}^{(0)}$, number of iterations $N$
\State \textbf{Output:} A Markov chain $\{\boldsymbol{\delta}^{(t)}\}_{t=0}^{N}$ approx. $p(\boldsymbol{\delta}\mid\mathbf{X})$
\State Initialize $\boldsymbol{\delta}^{(0)}$
\For{$t = 1$ to $N$}
    \State Sample candidate $\boldsymbol{\delta}' \sim q(\boldsymbol{\delta}' \mid \boldsymbol{\delta}^{(t-1)})$
    \State Compute acceptance probability
    \[
    \alpha = \min\!\Biggl(1, \frac{p(\mathbf{X}\mid\boldsymbol{\delta}')p(\boldsymbol{\delta}')}{p(\mathbf{X}\mid\boldsymbol{\delta})p(\boldsymbol{\delta})}\Biggr)
    \]
    \State Draw $u \sim \text{Uniform}(0,1)$
    \If{$u \leq \alpha$}
        \State Accept: $\boldsymbol{\delta}^{(t)} \gets \boldsymbol{\delta}'$
    \Else
        \State Reject: $\boldsymbol{\delta}^{(t)} \gets \boldsymbol{\delta}^{(t-1)}$
    \EndIf
\EndFor
\State \textbf{return} $\{\boldsymbol{\delta}^{(t)}\}_{t=0}^{N}$
\end{algorithmic}
\end{algorithm}

\subsection{Selection and Definition of Prior Knowledge }
The choice of prior directly impacts the prediction results of Bayesian methods. Different prior types can be seen as competing hypotheses for their ability to model a system, and prior suitability is tested with prior predictive checks \cite{van2021bayesian}. Depending on the available information, the priors can range from non-informative (diffuse) to highly informative.
In this paper, three prior types simulate different available information levels: diffuse, informative and highly informative. The diffuse prior includes generally available information such as lower and upper bounds of MP, and serves as a baseline for our comparison. 
The informative prior is derived by empirical evaluation of the MP distributions of six 6-DOF robots as shown in Fig. \ref{empiricalprior}. This 'empirical' prior distribution is used, if robot-specific CAD data is unavailable.
The highly informative or CAD prior possesses the most information in form of robot-specific CAD-data.

\subsubsection{Physical feasibility}
Regardless of the prior type, our method ensures physical feasibility of the identified parameters by making the following standard assumptions \cite{gaz2016extracting, lee2019geometric}:
\begin{itemize}
    \item Mass $m$ and motor inertia $J$ are positive
    \item Diagonal components $I_{diag}$ of inertia tensor are positive. Inertia tensor $I$ fulfills triangle inequalities and is overall positive definite
    \item CoM $r_i$ is bounded by the link geometry
    \item Predicted inertia $X_i$ must be positive
\end{itemize}
Parameter positivity constraints are encoded using the PDF $p(\delta_i)$. For inertia tensors, multiple MPs need to be jointly constrained. Thus, we introduce a conditional probability, a child node to all tensor components of the link, in the Bayesian network. In Fig. \ref{fig:bayes_network}, node "$\Delta, \succ$" implements a constant and nonzero probability, where the constraints on the tensor are fulfilled or set to zero otherwise. Similarly, node "$>0$" is a conditional probability derived from all MP, to ensure an overall positive inertia. For CoM, a parametrization in spherical coordinates by $r_k, \theta_k, \varphi_k$ simplifies the description of bounded length and direction. 
\begin{table*}[ht]
\caption{Diffuse and empirical prior for the mechanical parameters of MABI Max 100 with distributions from 'Prior Definition'.}
\label{priors}
\begin{center}
\renewcommand{\arraystretch}{1.1}
\begin{tabular}{|cc||l|l|l|l|l|}
\hline
\multicolumn{1}{|c|}{}                                                                                  & Body & \multicolumn{1}{c|}{$m$ {[}$kg${]}} & \multicolumn{1}{c|}{$r$ {[}$m${]}} & \multicolumn{1}{c|}{$J$ {[}$kg\cdot m^2${]}} & \multicolumn{1}{c|}{$I_{diag}$ {[}$kg\cdot m^2${]}} & \multicolumn{1}{c|}{$I_{off}$ {[}$kg\cdot m^2${]}} \\ \hline\hline
\multicolumn{1}{|c|}{\multirow{6}{*}{\rotatebox[origin=c]{90}{\begin{tabular}[c]{@{}c@{}}Empirical\\ Distribution\end{tabular}}}} & 1    & $\mu$ = 344.92, $\sigma$ = 138.55   & $\mu$ = 0.2388, $\sigma$ = 0.0404  & $\mu$ = 217.160, $\sigma$ = 217.16            & $\mu$ = 0, $\sigma$ = 20.771                        & $\mu$ = 0, $\sigma$ = 32.968                        \\ \cline{2-7} 
\multicolumn{1}{|c|}{}                                                                                  & 2    & $\mu$ =121.57 , $\sigma$ = 80.92    & $\mu$ = 0.4824, $\sigma$ = 0.0670  & $\mu$ = 635.076, $\sigma$ = 635.08           & $\mu$ = 0, $\sigma$ = 16.158                        & $\mu$ = 0, $\sigma$ = 18.388                       \\ \cline{2-7} 
\multicolumn{1}{|c|}{}                                                                                  & 3    & $\mu$ = 153.91, $\sigma$ = 86.77    & $\mu$ = 0.1261, $\sigma$ = 0.0606  & $\mu$ = 86.192, $\sigma$ = 86.192            & $\mu$ = 0, $\sigma$ = 97.039                        & $\mu$ = 0, $\sigma$ = 0.3628                       \\ \cline{2-7} 
\multicolumn{1}{|c|}{}                                                                                  & 4    & $\mu$ = 55.47, $\sigma$ = 40.28     & $\mu$ = 0.5393, $\sigma$ = 0.3040  & $\mu$ = 29.752, $\sigma$ = 29.752            & $\mu$ = 0, $\sigma$ = 23.289                        & $\mu$ = 0, $\sigma$ = 0.0660                       \\ \cline{2-7} 
\multicolumn{1}{|c|}{}                                                                                  & 5    & $\mu$ = 26.27, $\sigma$ = 15.24     & $\mu$ = 0.0787, $\sigma$ = 0.0437  & $\mu$ = 14.580, $\sigma$ =  14.580           & $\mu$ = 0, $\sigma$ = 0.2113                        & $\mu$ = 0, $\sigma$ = 0.0101                       \\ \cline{2-7} 
\multicolumn{1}{|c|}{}                                                                                  & 6    & $\mu$ = 25.99, $\sigma$ = 26.62     & $\mu$ = 0.2293, $\sigma$ = 0.2309  & $\mu$ = 6.362, $\sigma$ = 63.620             & $\mu$ = 0, $\sigma$ = 0.2801                        & $\mu$ = 0, $\sigma$ = 0.0041                       \\ \hline\hline
\multicolumn{2}{|c||}{Diffuse Bounds}                                                                           & $a$ = 0, $b$ = 1050           & $a$ = -1, $b$ = 1            & $a$ = $J_i$, $b$ = $2J_i$              & $a$ = 0, $b$ = 50                             & $a$ = -2, $b$ = 2                            \\ \hline
\end{tabular}
\end{center}
\end{table*}

\subsubsection{Diffuse Prior}
The constraints described in the last section ensure physical feasibility, but only partially specify the parameter intervals for the identification process. Therefore, we define a diffuse prior with numerical bounds for the MP of the MABI Max 100 in Table \ref{priors} using publicly available data from datasheets for total mass and rotor inertia. The prior predictive capability highly depends on the chosen interval, therefore a stark over- or underestimation is not desired:
\begin{itemize}
    \item Each link mass is bounded by the datasheet total mass
    \item Inertia tensor entries are bounded by Steiner's theorem
    \item Rotor inertia is lower than assembled motor inertia
    \item Axis total inertia is larger than the motor inertia
    \item Axis total inertia can be bounded by Steiner's theorem
\end{itemize}
Apart from hard bounds, conditional probabilities or effective coordinate choices can impart additional probabilistic information such as: 
\begin{itemize}
    \item Sum of base and body masses is close to the total mass
    \item CoM and translation vector directions are likely aligned 
    \item Inertia tensor principle axes might align with translation
\end{itemize}
These soft constraints add on to the 'diffuse' prior, showcasing the incremental benefit of richer information as our 'informed diffuse' prior, which is evaluated in Section IV. The interval limits established for the diffuse prior carry over to the empirical prior discussed next. 

\subsubsection{Empirical Prior}
\begin{figure}[ht]
    \centering
    \includegraphics[width=0.5\textwidth]{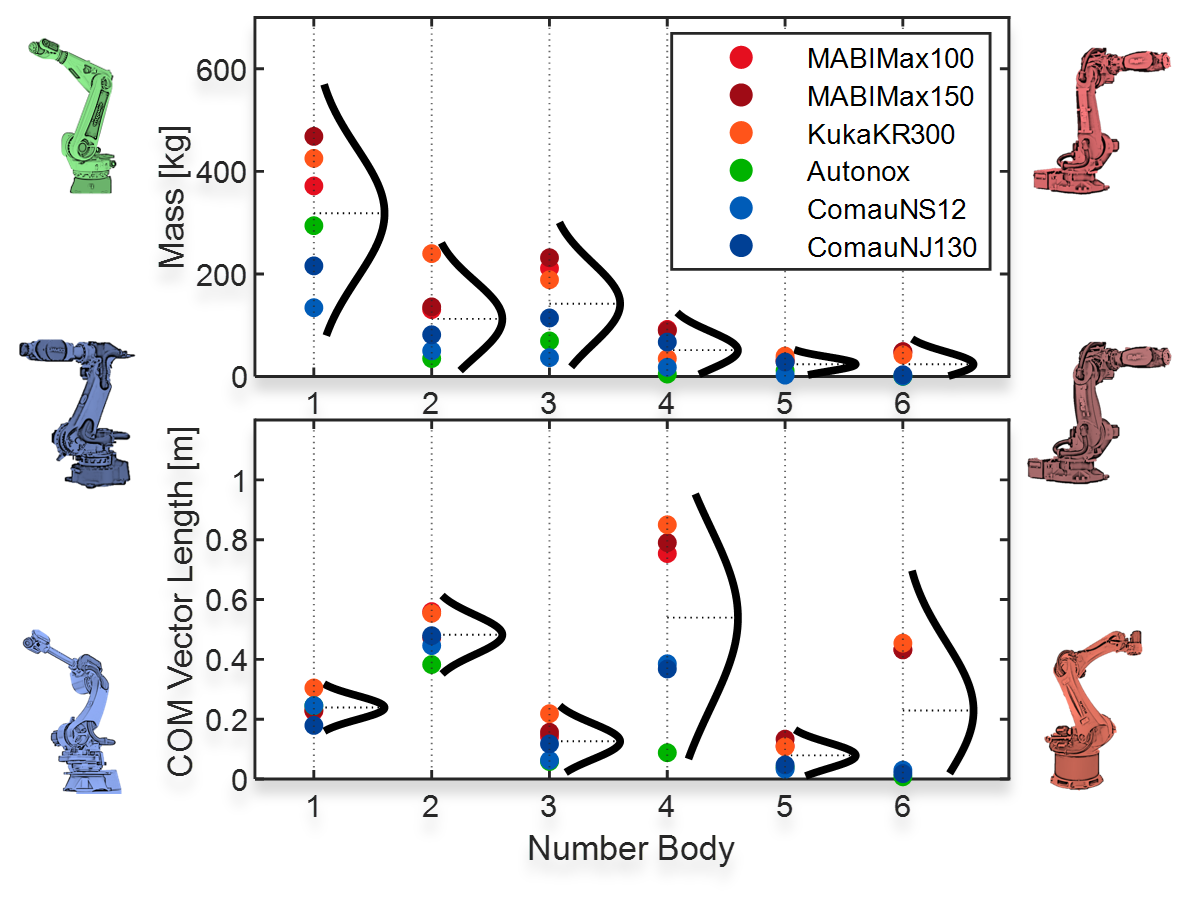}
    \caption{Evaluation of empirical prior distribution (unscaled) for mass and center of mass (CoM) length for axes 1 to 6 using six industrial robots.}
    \label{empiricalprior}
\end{figure}

For the empirical prior, we analyzed the CAD models of six industrial 6-DOF robots: MABI Max 100, MABI Max 150, Kuka KR300, Autonox AT00005, Comau NS12 and Comau NJ130 as shown in Fig. \ref{empiricalprior}. The datasheet total masses range from 335 kg (Comau NS12) to 2150 kg (Kuka KR300).
From the CAD models, we extract the following data points for each axis: link mass, length of CoM vector, diagonal- and off-diagonal components of the inertia tensor. For a better adaption to the MABI Max 100, the robot used in Section IV for evaluation of our method, the masses were scaled up by the ratio of its datasheet mass of 1050 kg to the average datasheet mass of 969 kg. The gathered and scaled data points are used to identify prior distribution parameters for the MABI Max 100, which are shown in Table \ref{priors}. We acknowledge the bias introduced by not excluding the Max 100 from the prior data, which we found would increase MAEs by up to $1\%$ for the 'empirical' prior results. Since the error is small, we opted for allowing a larger empirical sample size instead. The motor inertia distribution is unknown and a standard deviation equal to the nominal value is assumed. The CoM length distribution describes the radius in a spherical parametrization for the CoM. We can sample CoM from a cone centered on the translation vector, by choosing the z-axis as coinciding with the translation vector. The polar angle distribution then allows to directly control the probable angle between translation and CoM vector. For the empirical prior, we limit the polar angle to 90°, while the azimuth angle is unconstrained. As a side effect, the Metropolis algorithm appears to sample CoM more efficiently for these coordinates.

\subsubsection{CAD Prior}
If robot specific CAD data is available, a good initial model for the MP is provided for and the task is to update this model to the real system using measurements. Here, we choose the prior mean according to the CAD values and a variance of 10\% to the mean.

\subsubsection{Prior Definition}
The choice of prior distribution to model prior knowledge depends on available information, with many standard options available \cite{gelman1995bayesian}. Since the diffuse prior only considers lower and upper bounds $a$ and $b$, a uniform distribution is sufficient:
\begin{equation}
    \delta_i \sim \mathrm{Uniform}(a,b)
\end{equation}
For the CAD prior, normal distributions are used, since a good estimate of the parameters' mean $\mu$ is known and variances $\sigma^2$ are small enough to ensure physical feasibility. For the empirical prior, we only fit a zero mean normal distribution to the off-diagonal tensor components.
\begin{equation}
    \delta_i \sim \mathcal{N}(\mu, \sigma^2)    
\end{equation}
The half-normal or more generally, truncated normal distribution, is the foundation of the empirical prior, combining physical feasibility bounds $a,b$ with statistical information $\mu, \sigma$. In our case, it is used for mass, motor inertia, CoM length (i.e. radius) and the measurement noise estimate:
\begin{equation}
\delta_i \sim \mathrm{TruncNormal}(\mu, \sigma^2, a, b)    
\end{equation}
A log-normal distribution is used for diagonal tensor components in the empirical model, as they are guaranteed to be positive, but may vary greatly in magnitude:
\begin{equation}
    \delta_i \sim \mathrm{Lognormal}(\mu, \sigma^2)
\end{equation}

\subsection{Zero-Shot vs. Inference Mode}
The resulting method is bijective since it is able to learn the pose-dependent inertia of a robot but also determine the MP with statistical verification tools in each stage. In the prior prediction or Zero-Shot mode, the method works without additional measurements. This is inspired by the ZSL approach, an ML concept that does not require training data for learning \cite{NIPS2013_2d6cc4b2}. Although most ZSL methods have been applied to image classification problems using e.g. Bayesian methods, others have used them for regression problems as well \cite{zsl19}. 
In the backward or inference mode, the method uses experimental measurements to inform its learning process. For data generation, we use periodic excitation trajectories and a simplified elastic joint model \cite{siciliano2008springer} following the strategy presented in \cite{grundel2024modellierungsmethoden}.

\section{RESULTS AND DISCUSSION}
\label{results}

Prior predictive (Zero-Shot) abilities and inferred posterior results for the MABI Max 100 are evaluated for MP, BP and pose-dependent inertia. The evaluations in Section III.A. to D. focus on results for the empirical prior, as it is precise in prior and posterior prediction, without requiring robot-specific CAD data. In Section III.E., the results for all prior types are discussed. The diffuse prior is our baseline for the Bayesian approach. We compare it to a global optimization in MP space with bounds equivalent to the diffuse prior, using the Matlab function 'fmincon' and multiple restarts similar to \cite{xu2020dynamic}. Both inference and global optimization were run overnight for 8 hours on an Intel i5-1135G7. The measurement data consists of 147 configurations and associated point estimates for the inertia. 75 poses were randomly chosen for training and the remaining 72 for testing. The training data observation matrix is well conditioned with $\mathrm{cond}(\textbf{Y)}=33.13$ allowing an LS solution in BP and inertia space, indicating a potential excitation of all MP.

For evaluation of the Bayesian method for recovering MP, a mean absolute error (MAE) is selected and expressed as a percentage error compared to the nominal CAD model of the MABI Max 100. Compared to the root mean square error (RMSE), the MAE allows for equal weighting of errors \cite{Verma2020}.
\begin{equation}
    MAE=\frac{\overline{\bf{A}-\bf{B}}}{\overline{\bf{B}}}*100\%
\end{equation}
Since the CAD model does not contain the gear inertia, wires and attachments to the robot, some parameters are expected to be smaller than their real value. Therefore, a measure for congruency independent of scale is needed. For this purpose, we choose the cosine similarity, defined as \cite{Verma2020}:
\begin{equation}
    CS=\cos(\alpha)=\frac{\bf{<A,B>}}{||\bf{A}||\,||\bf{B}||}\in[-1,1]
\end{equation}
In the inertia space, limitation of the maximum errors and comparability to traditional fittings is essential. Therefore, we choose the RMSE and express it as a percentage relative to the measured inertia (RMSE\%).
For probabilistic comparison of sampled prior and posterior, we want to quantify the impact of the learning step. The desired result is a reduced posterior variance and the distribution mean shifting closer to the nominal parameter value. A change in mean or variance is also measureable as a decrease in the overlap of prior and posterior distribution. With the total variation distance (TVD) expressed using the $L^1$ distance, a simple measure for the similarity of PDF is given \cite{devroye2023totalvariationdistancehighdimensional}. 
\begin{equation}
    TVD_i = \frac{1}{2}\sum_{\delta_i}|p(\delta_i\mid \mathbf{X})-p(\delta_i)|\in [0,1]
\end{equation}
\subsection{Prior vs. Posterior}
\begin{figure}[ht]
    \centering
    \includegraphics[width=0.45\textwidth]{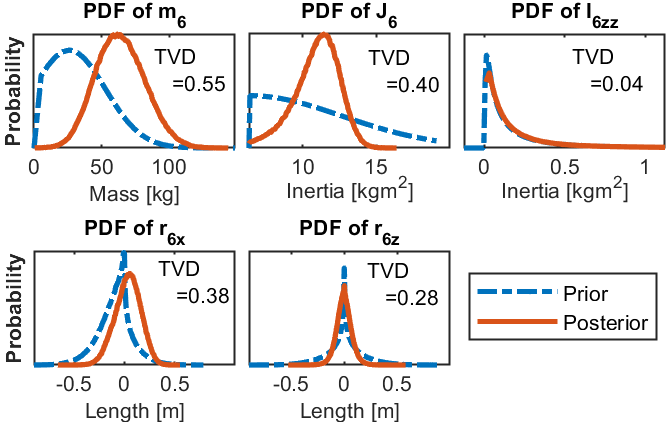}
    \caption{Prior and posterior ('Empirical') probability density functions (PDF) with total variation distance (TVD) of the mechanical parameters of axis 6.}
    \label{fig:PriorPosteriorPDFsAX6}
\end{figure}

Each inferred MP and the inferred measurement variance possess prior and posterior distributions with changes quantifiable by the TVD. For the sake of brevity, we focus on the MP describing BP \eqref{eq:BP_AX6}. The PDF and TVD are shown in Fig. \ref{fig:PriorPosteriorPDFsAX6}. For $m_6$, prior and posterior show a large TVD reflecting the visible change in distribution mean. The posterior reveals that a mass close to zero is improbable given the measured inertia. The motor inertia $J_6$ also displays a significant TVD and learning in terms of a reduction in variance. The posterior for the inertia tensor component $I_{6zz}$ on the other hand mostly coincides with the prior and the low TVD matches this observation of minor learning. 
In general, CoM and masses show an increase in TVD with each successive body. This variability in learning success can be attributed to the growing distance between bodies and axis one as well as the resulting increase of Steiner components. Diagonal and off-diagonal tensor components exhibit the lowest change and appear strongly predetermined by the prior choice. 

\subsection{Learning Mechanical Parameters}
\begin{figure}[ht]
    \centering
    \includegraphics[width=0.46\textwidth]{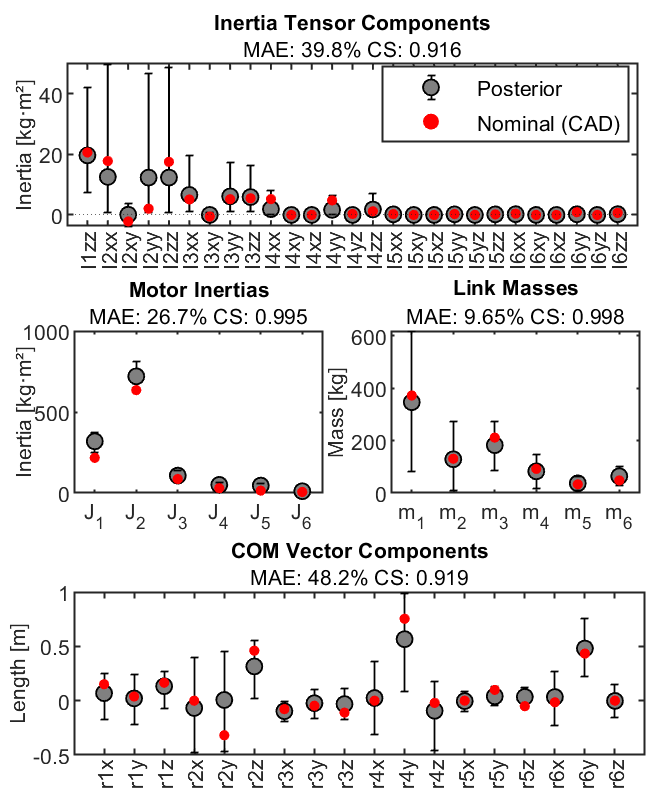}
    \caption{Posterior distributions ('Empirical') for the mechanical parameters inferred from 75 measurements with 95\% confidence intervals.}
    \label{fig:PosteriorMPempirical}
\end{figure}

For the MP, the expected values and variances of the empirical prior can be derived from Table \ref{priors}. Yet, lower and upper bounds are needed to parameterize the truncated normal or lognormal distributions in order to ensure physical feasibility.
The confidence intervals of the prior include the nominal CAD value, which confirms the choice of prior and serves as a prior predictive check. In Fig. \ref{fig:PosteriorMPempirical} the posterior update is shown, which further improves on the prior results. 

Here lies a main advantage of our method, since we can directly learn MP from measurements, which can be fed back into CAD-based simulation or RNEA calculations.
In contrast to linear matrix inequality and optimization-based approaches like fmincon, all learning parameters lie within the allowed space far from the optimization bounds. Masses are identified with the smallest MAE percentage, although variances in the lower axes are high. In contrast, the motor inertia shows a higher error, but smaller confidence intervals
Since the nominal CAD model does not include the gears, a higher learned value for the motor inertia was expected.
With the TVD being low for inertia tensor components, only minor reductions of error and small increases of cosine similarity are observed. Hence, an informative prior or additional constraints are necessary for successful identification. 

\begin{table*}[ht]
\caption{Evaluation of mean absolute error (MAE), cosine similarity (CS) and root mean squared error (RMSE) for the approaches.}
\label{evaluation}
\begin{center}
\renewcommand{\arraystretch}{1.05}
\begin{tabular}{|l| |rlrl|rlrl| |rlrl|}
\hline
Approach / Evaluation Metric     & \multicolumn{2}{c|}{MAE {[}\%{]}}                                        & \multicolumn{2}{c|}{CS}                           & \multicolumn{2}{c|}{MAE {[}\%{]}}                                       & \multicolumn{2}{c||}{CS}                           & \multicolumn{4}{c|}{RMSE {[}\%{]}}                                                                                 \\ \hline\hline
Ordinary Least Squares           & \multicolumn{4}{c|}{indeterminable}                                                                       & \multicolumn{2}{r|}{33.37}                                              & \multicolumn{2}{r||}{0.990}                        & \multicolumn{2}{r|}{5.30}                             & \multicolumn{2}{r|}{9.81}                         \\ \hline
Gradient-based (fmincon)         & \multicolumn{2}{r|}{142.64}                                              & \multicolumn{2}{r|}{0.524}                        & \multicolumn{2}{r|}{18.22}                                              & \multicolumn{2}{r||}{0.997}                        & \multicolumn{2}{r|}{6.18}                                      & \multicolumn{2}{r|}{7.84}                         \\ \hline
Diffuse (Prior / \underline{Posterior})      & \multicolumn{1}{r|}{189.83}  & \multicolumn{1}{r|}{{\underline{87.54}}}          & \multicolumn{1}{r|}{0.605} & {\underline{ 0.847}}          & \multicolumn{1}{r|}{486.62} & \multicolumn{1}{l|}{{\underline{ 26.42}}}          & \multicolumn{1}{r|}{0.979} & {\underline{ 0.991}}          & \multicolumn{1}{r|}{354.69} & \multicolumn{1}{l|}{{\underline{ 17.42}}} & \multicolumn{1}{r|}{340.86} & {\underline{ 19.23}}         \\ \hline
Diffuse inf. (Prior / \underline{Posterior}) & \multicolumn{1}{r|}{189.83}  & \multicolumn{1}{l|}{{\underline{ 83.12}}}          & \multicolumn{1}{r|}{0.789} & {\underline{ 0.895}}          & \multicolumn{1}{r|}{502.73} & \multicolumn{1}{l|}{{\underline{ 21.52}}}          & \multicolumn{1}{r|}{0.980} & {\underline{ 0.994}}          & \multicolumn{1}{r|}{366.14} & \multicolumn{1}{l|}{{\underline{ 13.19}}} & \multicolumn{1}{r|}{338.87} & {\underline{ 14.43}}         \\ \hline
Empirical (Prior / \underline{Posterior})    & \multicolumn{1}{r|}{47.98}   & \multicolumn{1}{l|}{{\underline{ \textbf{31.06}}}} & \multicolumn{1}{r|}{0.953} & {\underline{ \textbf{0.957}}} & \multicolumn{1}{r|}{31.24}  & \multicolumn{1}{l|}{{\underline{ \textbf{11.55}}}} & \multicolumn{1}{r|}{0.977} & {\underline{ \textbf{0.998}}} & \multicolumn{1}{r|}{23.06}  & \multicolumn{1}{r|}{{\underline{\textbf{7.27}}}}  & \multicolumn{1}{r|}{22.98}  & \multicolumn{1}{r|}{{\underline{ \textbf{7.65}}}} \\ \hline
CAD-Update (Prior / \underline{Posterior})    & \multicolumn{1}{r|}{18.79}   & \multicolumn{1}{l|}{{\underline{ 9.57}}} & \multicolumn{1}{r|}{0.999} & {\underline{ 0.998}} & \multicolumn{1}{r|}{19.44}  & \multicolumn{1}{l|}{{\underline{ 8.09}}} & \multicolumn{1}{r|}{0.993} & {\underline{ 0.999}} & \multicolumn{1}{r|}{23.29}  & \multicolumn{1}{r|}{{\underline{7.34}}}  & \multicolumn{1}{r|}{26.49}  & \multicolumn{1}{r|}{{\underline{ 7.36}}} \\ \hline\hline                   
\multicolumn{1}{|c||}{}           & \multicolumn{4}{c|}{Mechanical Parameters}                                                                                   & \multicolumn{4}{c||}{Base Parameters}                                                                                        & \multicolumn{2}{c|}{Inertia (Train)}                           & \multicolumn{2}{c|}{Inertia (Test)}                \\ \hline
\end{tabular}
\end{center}
\end{table*}

\subsection{Learning Base Parameters}

\begin{figure}[ht]
    \centering
    \includegraphics[width=0.44\textwidth]{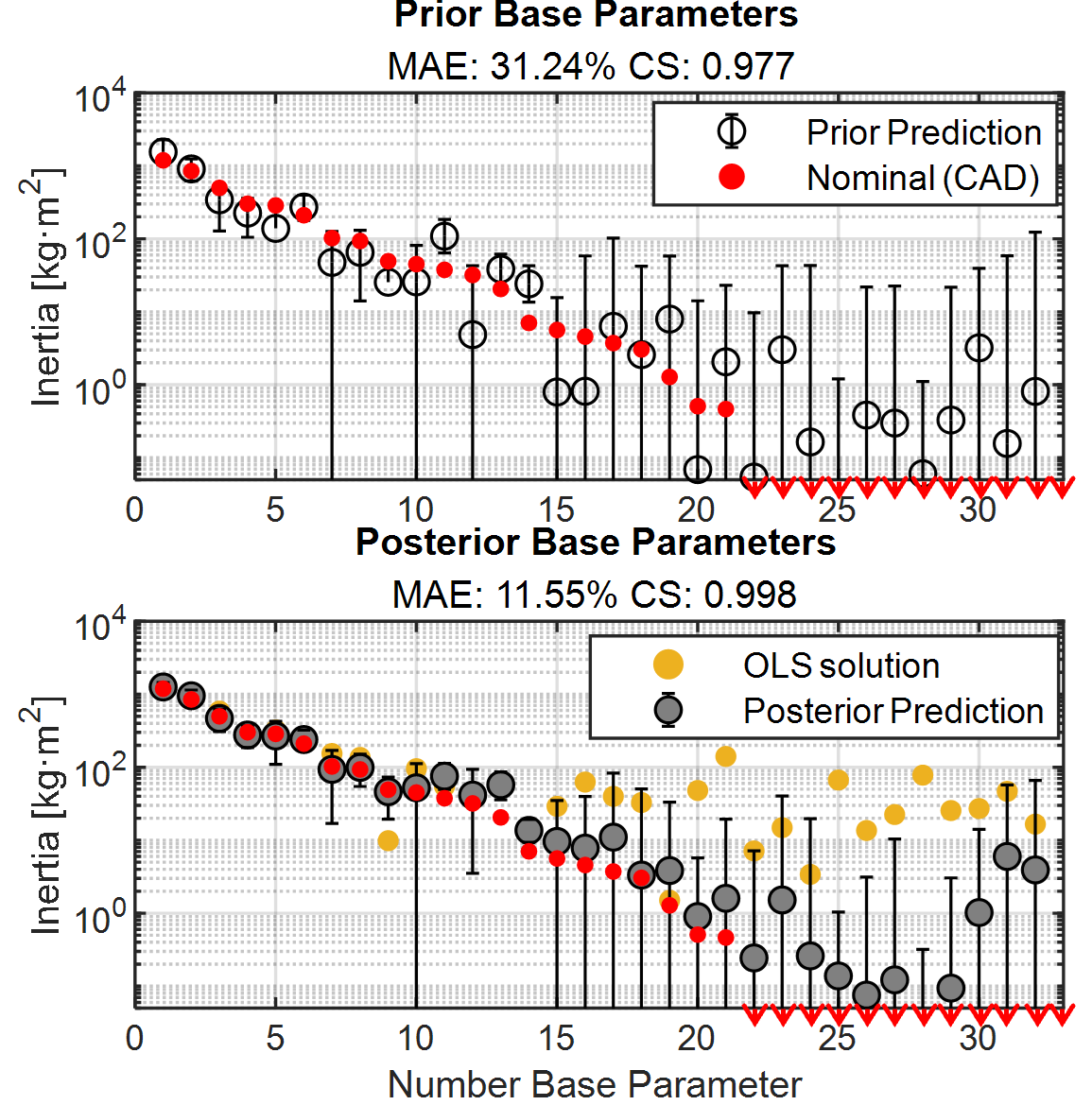}
    \caption{Logarithmic plot for prior and posterior distributions ('Empirical') of the 32 base parameters with 95\% confidence intervals.}
    \label{fig:PriorPosteriorBP}
\end{figure}

Fig. \ref{fig:PriorPosteriorBP} shows the learned BP of the MABI Max 100 using empirical prior, posterior and the state-of-the-art LS approach. Since the 32 BP vary greatly in size, a logarithmic plot of the absolute value is provided.

The parameters are ordered by descending value of the nominal CAD-based BP. An exponential decline in magnitude is visible. A prior predictive test is successful for 29 parameters with the nominal value lying within the predicted confidence interval. The test fails for the parameters containing the prior motor inertia of axis 4-6, since the information regarding the gears is missing.
Based on the empirical prior, the two parameters 25 and 28 are unidentifiable, as the confidence interval is confined to values smaller than $1\,kgm^2$ and therefore of similar magnitude as the measurement noise. The posterior shows a low MAE and a high cosine similarity with the nominal model. Hence, there is no indication that the inferred set of BP must be reduced to essential parameters as sometimes suggested for LS approaches.

\subsection{Learning Pose-dependent Inertia Parameters}

\begin{figure}[ht]
    \centering
    \includegraphics[width=0.48\textwidth]{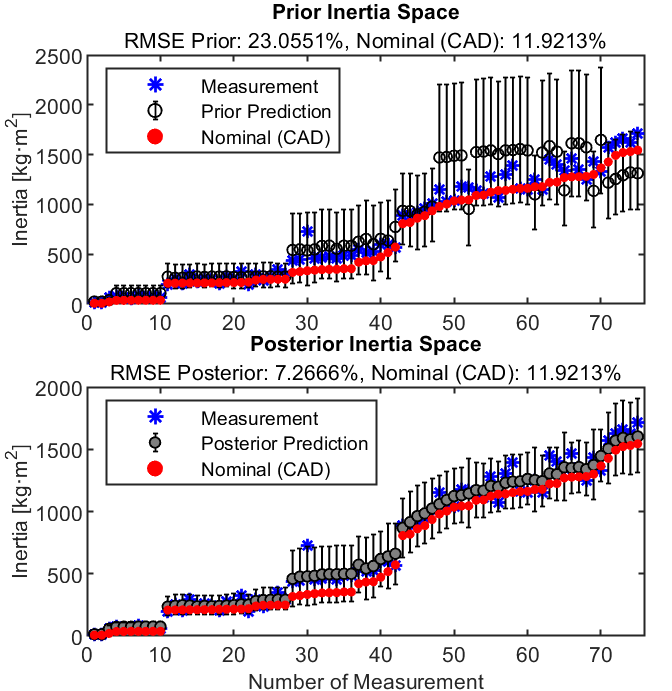}
    \caption{Prior and posterior predictions ('Empirical') with 95\% confidence intervals for 75 measurements. Data of all robot axes by ascending inertia.}
    \label{fig:PriorPosteriorInertiaModel}
\end{figure}

In inertia space, the empirical prior is able to reflect both nominal values as well as the measurements (Fig. \ref{fig:PriorPosteriorInertiaModel}). The posterior update provides smaller confidence intervals and a good match to the measurement data without overfitting associated with the LS approach.

\subsection{Discussion}
Table \ref{evaluation} provides a comparison of different priors and resulting posteriors after inference using the measured inertias. The MP MAE is given as the average of the four individual MAE of the MP, as can be seen in Fig. \ref{fig:PosteriorMPempirical}. In general, the more informative the prior, the closer prior and posterior are to the nominal model. As a result, the MAE percentage decreases and the cosine similarity increases.
While LS cannot provide a MP space solution, the gradient-based method can, but with large errors. The diffuse and informed diffuse approaches improve the MAE and CS, but are limited by the conservative choice of bounds, stressing the advantage of the empirical prior in providing a regularized sampling space while remaining interpretable. If an uncertain CAD model is the baseline, the results should not be interpreted as errors, but relative changes. Intuitively, the prior would have zero error as it is based on the nominal model and is also compared to it. Deviations are due to numerical errors, nonlinearity of the BP transform, and bias introduced by the truncated normal PDF of the motor inertia. At last, the posterior can be seen as the new reference model, which updates the CAD model to the real system.

\section{CONCLUSIONS}
We presented a Bayesian approach that incorporates priors as probability densities to regularize the dynamic parameter identification for industrial robots. By analyzing CAD data of six robots and defining various prior levels—ranging from diffuse to highly informative—results improve with additional information, yet remain physically feasible even without datasheets or CAD. With few measurements refining estimates, rank‐deficiency as in constraint work spaces is handled and confidence intervals are narrowed. We also show how to update CAD models experimentally. Future work includes validating these parameters in feed-forward control and exploring advanced sampling (e.g. Hamiltonian Monte Carlo). The resulting uncertainty bounds benefit robust control, safety, and supervisory applications.

\addtolength{\textheight}{-12cm}   





\section*{ACKNOWLEDGMENT}

We gratefully acknowledge the support of the Siemens AG and the support by D. Bitterolf and S. Tauchmann. We used OpenAI's ChatGPT o1 to develop and format the paper.




\bibliographystyle{IEEEtran}
\bibliography{IEEEexample}



\end{document}